\documentclass{article}
\usepackage{codefuse_tech_report}
\usepackage[colorlinks = true,
            linkcolor = blue,
            urlcolor  = blue,
            citecolor = blue,
            anchorcolor = blue]{hyperref}   
\usepackage{microtype}
\usepackage{xurl}
\usepackage{booktabs}
\usepackage{enumitem}
\usepackage{multicol}
\usepackage{CJKutf8}
\usepackage{siunitx}
\usepackage{floatflt}
\usepackage{graphicx}
\usepackage{wrapfig}
\usepackage{authblk}
\usepackage{lipsum}
\usepackage{algorithm}
\usepackage{algorithmicx}
\usepackage{algpseudocode}
\usepackage{subfigure}
\usepackage{multirow}
\usepackage{pifont}  

\newcommand{\tb}[1]{\textbf{#1}}

\algnewcommand{\LeftComment}[1]{\Statex \(\triangleright\) #1}

\usepackage{array}
\usepackage{amsmath}
\usepackage{amssymb}
\usepackage{mathtools}
\usepackage{amsthm}

\usepackage[capitalize,noabbrev]{cleveref}
\usepackage{adjustbox} 

\theoremstyle{plain}

\theoremstyle{definition}

\theoremstyle{remark}

\colmfinalcopy

\usepackage[utf8]{inputenc}
\usepackage[T1]{fontenc}
\usepackage{caption} 
\usepackage{adjustbox} 
\usepackage{fontawesome5}

\usepackage{tcolorbox}
\tcbuselibrary{skins,breakable}

\tcbuselibrary{skins}

\usepackage{color}
\usepackage[table]{xcolor}
\usepackage{soul} 

\definecolor{tred}{RGB}{251, 130, 132}
\definecolor{torange}{RGB}{247, 162, 116}
\definecolor{tyellow}{RGB}{251, 218, 140}
\definecolor{tgreen}{RGB}{127, 204, 181}
\definecolor{tblue}{RGB}{89, 177, 215}
\definecolor{insightblue}{RGB}{162, 210, 255}
\definecolor{questionred}{RGB}{255, 175, 204}

\title{\centering F2LLM-v2: Inclusive, Performant, and Efficient Embeddings\\for a Multilingual World}

\author{%
Ziyin Zhang$^{1,2}$
~~Zihan Liao$^{1}$
\\

\vspace{-6pt}
\bf
~~Hang Yu\thanks{Correspondence to: Hang Yu \textless hyu.hugo@antgroup.com\textgreater, Peng Di \textless dipeng.dp@antgroup.com\textgreater, Rui Wang \textless wangrui12@sjtu.edu.cn\textgreater.}~~$^{,1}$ 
~~Peng Di$^{*, 1}$
~~Rui Wang$^{*, 2}$

\vspace{10pt}
$^1$Ant Group\ \ \ $^2$Shanghai Jiao Tong University\\
\vspace{10pt}
\hspace{-5.5pt}~~~\faGithub ~~\href{https://github.com/codefuse-ai/CodeFuse-Embeddings/tree/main/F2LLM}{github.com/codefuse-ai/CodeFuse-Embeddings}\\
\hspace{-8pt}\includegraphics[width=1em,height=1em]{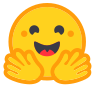} ~\href{https://huggingface.co/collections/codefuse-ai/f2llm}{huggingface.co/collections/codefuse-ai/f2llm}\\
}

\colmfinalcopy 

\begin{document}

\maketitle

\begin{figure}[h!]
    \centering
    \includegraphics[width=1\linewidth]{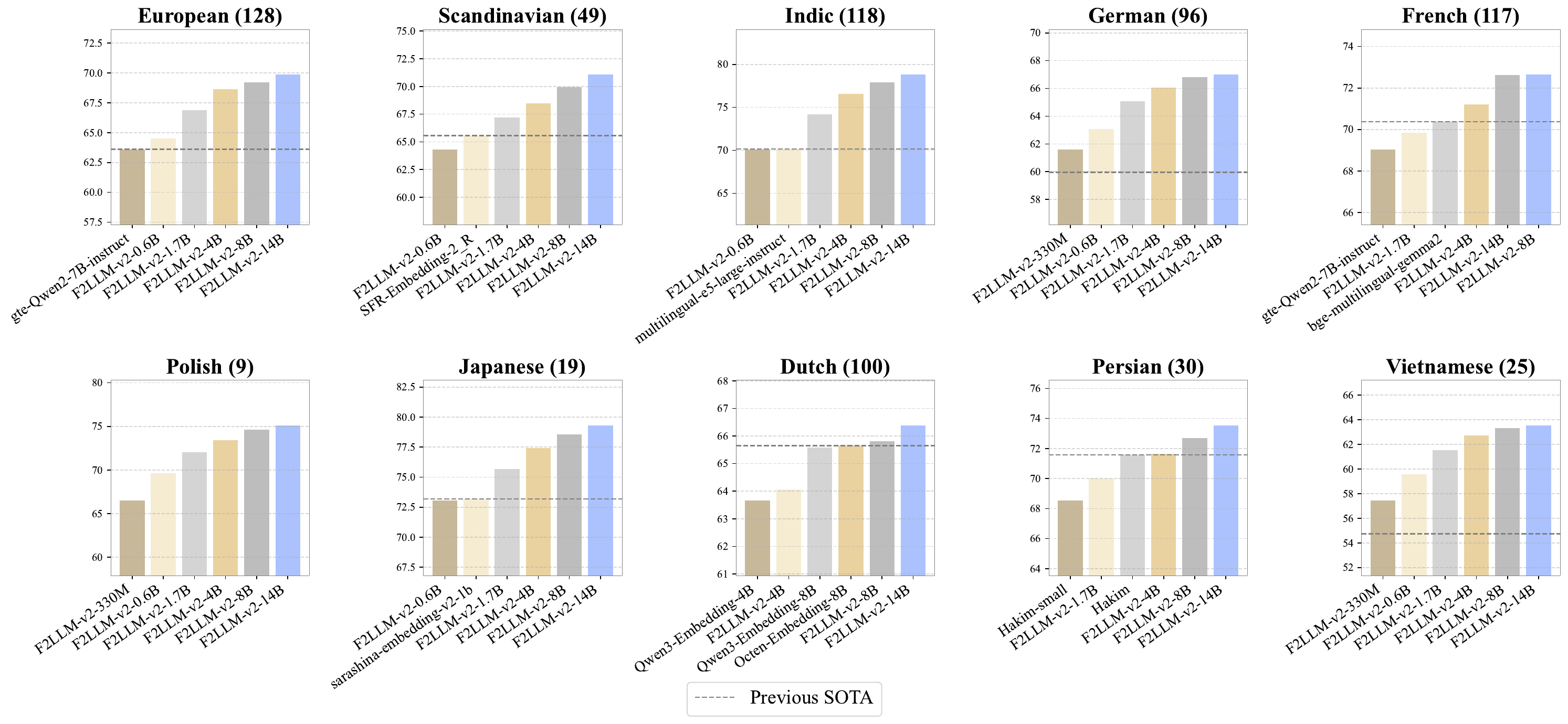}
    \caption{The top six models on ten language-specific MTEB leaderboards. The previous SOTA performance is given by the horizontal line. In each subplot title, we list the number of submissions with complete results on the corresponding benchmark. For comparison, the English benchmark has 163 complete submissions.}
    \label{fig:cover}
\end{figure}

\begin{abstract}
We present F2LLM-v2, a new family of general-purpose, multilingual embedding models in 8 distinct sizes ranging from 80M to 14B. Trained on a newly curated composite of 60 million publicly available high-quality data samples, F2LLM-v2 supports more than 200 languages, with a particular emphasis on previously underserved mid- and low-resource languages. By integrating a two-stage LLM-based embedding training pipeline with matryoshka learning, model pruning, and knowledge distillation techniques, we present models that are far more efficient than previous LLM-based embedding models while retaining competitive performances. Extensive evaluations confirm that F2LLM-v2-14B ranks first on 11 MTEB benchmarks, while the smaller models in the family also set a new state of the art for resource-constrained applications. To facilitate open-source embedding model research, we release all models, data, code, and intermediate checkpoints.
\end{abstract}

\section{Introduction}

Text embedding models serve as the fundamental backbone for a wide array of AI applications, including semantic search, retrieval-augmented generation (RAG), text classification, and clustering. By mapping unstructured text into dense vector spaces, these models allow machines to capture complex semantic relationships, enabling efficient and accurate information retrieval and data analysis across massive datasets. This field has recently transitioned from encoder-based architectures~\citep{2019BERT,2019RoBERTa,2020XLM-R} to decoder-based LLM embeddings~\citep{2025Qwen3-Embedding,2025NV-Embed,2025f2llm}, benefiting from the extensive reasoning and linguistic capabilities acquired during large-scale pre-training and achieving remarkable gains in performance.

Despite these advancements, the current state of frontier embedding research is characterized by two significant limitations. First, there is a pervasive English-centric bias in both model training and benchmark evaluation. While benchmarks such as MTEB have been instrumental in standardizing evaluation, the high-resource language subsets therein - such as English and Chinese - receive a disproportionately large share of attention, resulting in an abundance of models that are performant in English but fail to provide global utility. Second, a transparency gap has emerged within the research community. Most top-performing embedding models, such as Gemini-Embedding~\citep{2025Gemini-Embedding} and Qwen3-Embedding~\citep{2025Qwen3-Embedding}, are released either as closed-source APIs or open-weight models without disclosing the underlying training data or methodologies. This lack of transparency hinders reproducibility and limits our collective understanding of how to build truly inclusive, general-purpose embedding systems.

To directly tackle these challenges, we introduce F2LLM-v2, a new family of general-purpose, multilingual embedding models designed to address these critical imbalances. We curate a massive, high-quality training corpus of 60 million samples spanning 282 natural languages and over 40 programming languages solely from publicly available resources. By prioritizing real-world data availability over benchmark-specific optimization, we create a model family that excels across a truly global range of applications, including those involving underserved languages. Besides linguistic inclusivity, we also address computational inclusivity by providing 8 distinct model sizes, ranging from 80M to 14B parameters. By integrating Matryoshka Representation Learning (MRL) and a two-stage training pipeline enhanced by model pruning and novel knowledge distillation, we ensure high performance even in resource-constrained environments. Extensive evaluations confirm that our 14B model achieves state-of-the-art results on 11 MTEB benchmarks, setting a new standard for multilingual embedding capabilities, while the smaller models also outperform previous frontier models with a similar size. To foster an open and equitable research environment, we release the complete training recipe, intermediate checkpoints, and all associated code and data for the F2LLM-v2 family, aiming to drive progress toward a more inclusive future for AI technology.
\section{Related Work}

The previous generation of encoder-based embedding models witnessed a proliferation of massively multilingual embedding models supporting hundreds of languages, represented by XLM-R~\citep{2020XLM-R}, mDeBERTaV3~\citep{2023DebertaV3}, mBART~\citep{2020mBART}, and mT5~\citep{2021mT5}. Recently, decoder-based embedding models have become the dominant paradigm, benefiting from their extensive capabilities acquired during large-scale pre-training, as verified by state-of-the-art models such as E5-Mistral~\citep{2024E5-Mistral}, NV-Embed~\citep{2025NV-Embed}, Qwen3-Embedding~\citep{2025Qwen3-Embedding}, and Gemini-Embedding~\citep{2025Gemini-Embedding}.

However, this advancement has been accompanied by a shift toward English-centric evaluation. This is evidenced in MTEB~\citep{2023MTEB}, which has been established as one of the most recognized text embedding benchmarks, covering over 500 evaluation tasks and more than 250 languages~\citep{2025MMTEB}. Yet, in reality, the MTEB leaderboards exhibit significant linguistic bias. For instance, in the MTEB-Multilingual benchmark, 35 out of the 131 tasks focus exclusively on English, potentially obscuring a model's true multilingual efficacy. Furthermore, many language-specific benchmarks receive disproportionately less attention compared with the English or Multilingual benchmarks. As an extreme example, the Polish MTEB benchmark had only a single model with complete results before our models were submitted.

This disparity is exacerbated by the fact that many top-performing multilingual embedding models - such as Qwen3-Embedding~\citep{2025Qwen3-Embedding}, Gemini-Embedding~\citep{2025Gemini-Embedding}, and EmbeddingGemma~\citep{2025EmbeddingGemma} - are either closed-source APIs or open-weight only without training transparency. KaLM-Embedding~\citep{2025KaLM-Embedding-V2} represents one of the few exceptions with transparency in training data, but focuses exclusively on the Multilingual leaderboard and is not evaluated on the aforementioned language-specific benchmarks that are critical for truly global applications.
\section{F2LLM-v2}

\begin{figure}[t]
    \centering
    \includegraphics[width=1.0\linewidth]{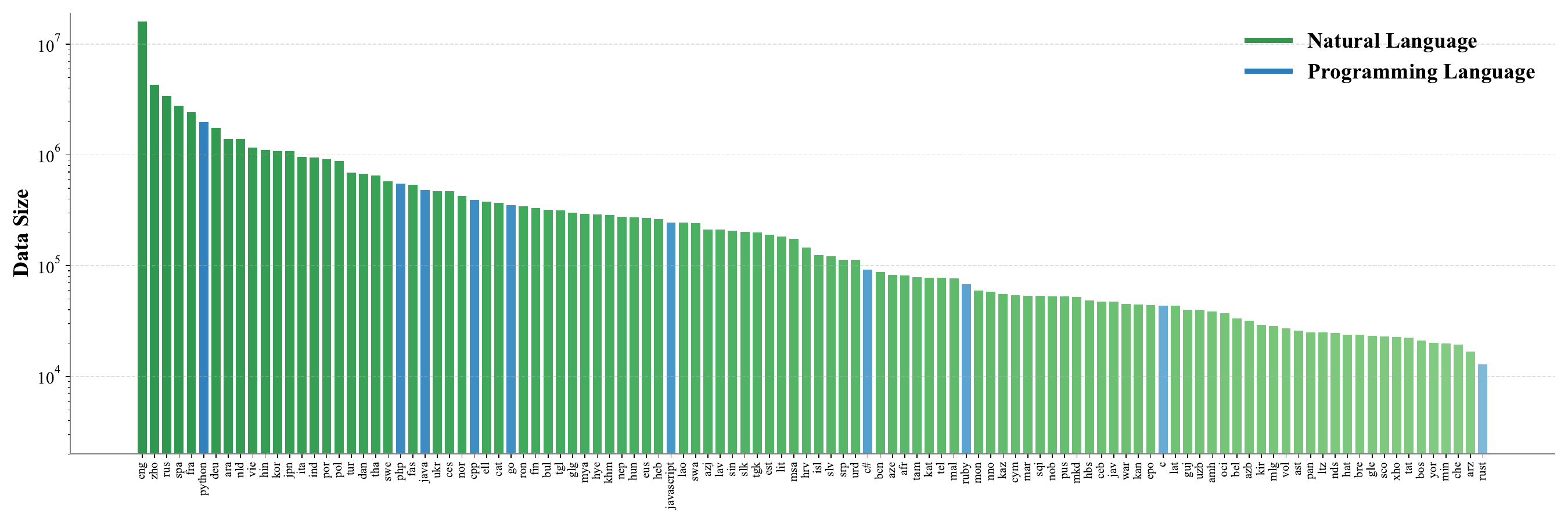}
    \caption{Top-100 natural languages and top-10 programming languages in our training data.}
    \label{fig:language-distribution}
\end{figure}

\subsection{Training Data}

\begin{wrapfigure}{r}{0.5\linewidth}
    \centering
    \includegraphics[width=1\linewidth]{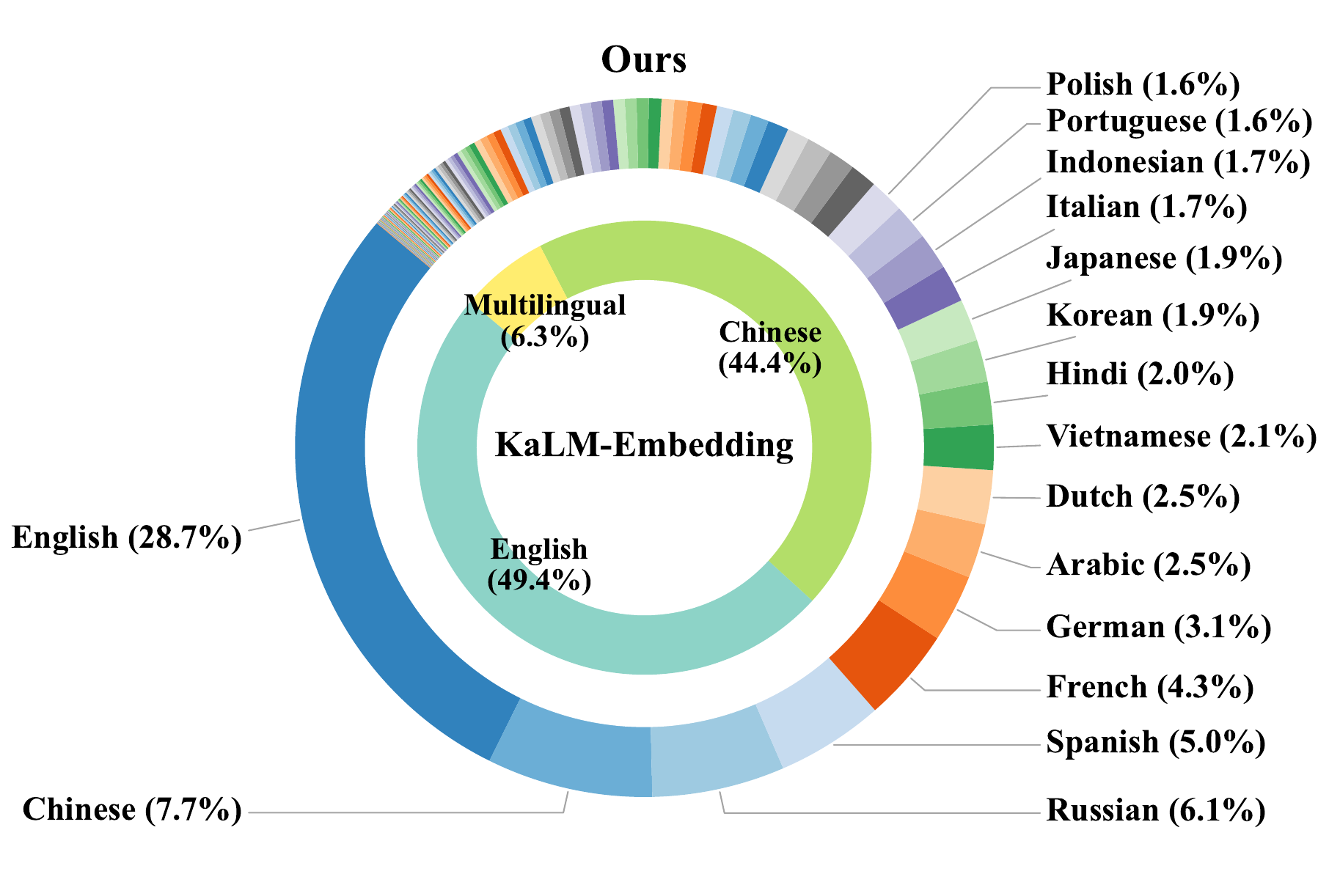}
    \caption{Comparison between the language distribution of our training data (outer circle) and KaLM-Embedding (inner circle). KaLM-Embedding's data is only annotated with three labels, while ours are annotated with specific languages.}
    \label{fig:language-comparison}
\end{wrapfigure}

A cornerstone of F2LLM-v2 is the compilation of a vast and diverse training corpus designed to foster both linguistic inclusivity and broad task competency. We aggregate data from 157 publicly available sources, creating a collection of 60 million training samples that span 282 natural languages (as identified by ISO-639-3 codes) and over 40 programming languages. Crucially, our data curation process is driven by real-world data availability rather than optimizing for specific benchmarks. For instance, our dataset contains substantial data for Spanish, Arabic, Italian, Indonesian, and Portuguese (Figure~\ref{fig:language-distribution}), despite these languages lacking dedicated benchmarks in MTEB. This approach, which also includes a long tail of low-resource languages and a significant volume of code, aims to build a model with truly global utility and stands in direct contrast to recent open-source datasets such as the one released by KaLM-Embedding~\citep{2025KaLM-Embedding-V2}, which is heavily skewed towards English and Chinese (Figure~\ref{fig:language-comparison}). We provide a more comprehensive linguistic breakdown of our dataset in Appendix~\ref{appendix:data}.

\begin{figure}
    \centering
    \includegraphics[width=0.7\linewidth]{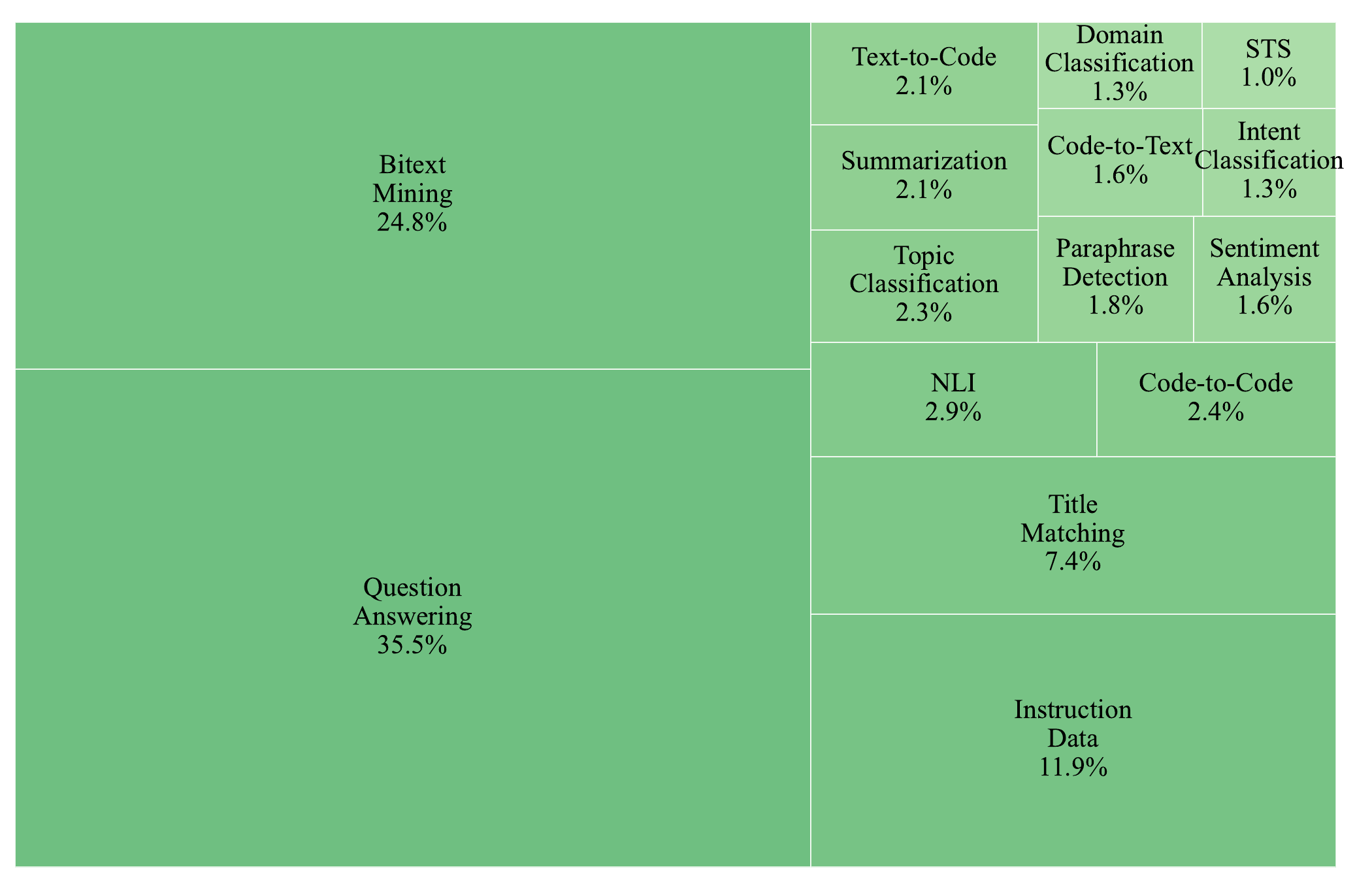}
    \caption{Task type distribution in our training data.}
    \label{fig:task-distribution}
\end{figure}

The functional diversity of our dataset is equally critical for training a general-purpose embedding model. As shown in Figure~\ref{fig:task-distribution}, our collection encompasses a wide spectrum of tasks, ranging from retrieval-focused question answering and bitext mining to classification-oriented sentiment analysis and intent/domain classification.

To leverage this heterogeneity within a unified contrastive learning framework, we follow the first generation of F2LLM~\citep{2025f2llm} and consolidate all data into three canonical formats: \emph{retrieval, clustering, and two-way classification}. This consolidation allows the model to learn a versatile embedding space by optimizing a single, consistent objective across disparate data sources and task structures. For the retrieval format, data consists of (query, positive document, hard negatives) tuples. We leverage both in-batch negatives, where other documents in a mini-batch serve as negatives, and explicitly provided hard negatives (mined using Qwen3-Embedding-8B) to create a challenging and efficient training signal. For the clustering format, which also ingests multi-class classification tasks, tuples are formed by sampling an anchor, a positive example from the same class, and a hard negative from a different class. Finally, the two-way classification format directly uses class labels, where a given text serves as the anchor, the corresponding label text is the positive, and the opposite label text is the negative. For both clustering and classification, only hard negatives are utilized to avoid introducing false negatives from in-batch samples.

\begin{table}[]
    \small
    \centering
    \begin{tabular}{lrrrrrrrr}
    \toprule
        & \tb{80M} & \tb{160M} & \tb{330M} & \tb{0.6B} & \tb{1.7B} & \tb{4B} & \tb{8B} & \tb{14B} \\
    \midrule
        \rowcolor{gray!15} \multicolumn{9}{c}{\textit{Model Configuration}}\\
        Hidden Size & 320 & 640 & 896 & 1024 & 2048 & 2560 & 4096 & 5120 \\
        MLP Intermediate Size & 2048 & 1536 & 2560 & 3072 & 6144 & 9728 & 12288 & 17408 \\
        Transformer Layers & 8 & 9 & 16 & 28 & 28 & 36 & 36 & 40 \\
        Attention Heads & 16 & 16 & 16 & 16 & 16 & 32 & 32 & 40 \\
        KV Heads & 8 & 8 & 8 & 8 & 8 & 8 & 8 & 8 \\
        Head Dimension & 128 & 128 & 128 & 128 & 128 & 128 & 128 & 128 \\
    \midrule
        \rowcolor{gray!15} \multicolumn{9}{c}{\textit{Model Size}}\\
        Embedding Parameters & 49M & 97M & 136M & 156M & 311M & 389M & 622M & 778M \\
        Non-Embedding Parameters & 31M & 62M & 198M & 440M & 1409M & 3634M & 6946M & 13212M \\
        Total Parameters & 80M & 159M & 334M & 596M & 1721M & 4022M & 7568M & 13990M \\
    \midrule
        \rowcolor{gray!15} \multicolumn{9}{c}{\textit{Training Configuration}}\\
        MRL Support & $\surd$ & $\surd$ & $\surd$ & $\surd$ & $\surd$ & $\surd$ & $\surd$ & $\surd$ \\
        Learning Rate & 4e-5 & 3e-5 & 2e-5 & 1e-5 & 9e-6 & 7e-6 & 6e-6 & 5e-6 \\
        Epochs & 4 & 3 & 3 & 2 & 2 & 2 & 2 & 2 \\
        Batch Size & 512 & 512 & 512 & 512 & 512 & 512 & 512 & 512 \\
        Teacher & 0.6B & 0.6B & 0.6B & 1.7B & 4B & - & - & -\\
    \bottomrule
    \end{tabular}
    \caption{F2LLM-v2 model and training configurations.}
    \label{tab:model_config}
\end{table}

\subsection{Model Architecture}

We train models in 8 distinct sizes: 80M, 160M, 330M, 0.6B, 1.7B, 4B, 8B, and 14B. All models adopt a standard dense Transformer decoder architecture based on Qwen3~\citep{2025Qwen3}, and utilize the final hidden states of the EOS token as sequence representation. The detailed model configurations are given in Table~\ref{tab:model_config}. Models from 0.6B to 14B directly correspond to Qwen3 LLMs, while the 80M, 160M, and 330M models are pruned from the 0.6B model.

\subsection{Two-stage Training}\label{sec:method-training}

We adopt a two-stage training strategy following previous works~\citep{2025NV-Embed,2025Qwen3-Embedding}. The first stage focuses on building a robust semantic foundation, and 7 retrieval datasets are selected based on their large scale and broad language coverage, totalling 27 million samples: CodeSearchNet, CodeSearchNet-CCR, OpenCodeGeneticInstruct, WebFAQ, MMARCO, CLIRMatrix, and ParaCrawl (refer to Appendix~\ref{appendix:data} for details). Five models (0.6B-14B) are trained in this stage, and we employ the raw data without applying any instructional prefix.

The second stage aims to sharpen the model's ability to handle the nuances of diverse downstream applications like classification, reranking, and paraphrase detection. For this stage, we sample at most 80 thousand queries from each data source, producing a mixture of 18 million samples. We apply task-specific instructions to the queries, and also randomly apply instructions to 30\% of documents and negatives in tasks where queries and documents are symmetric, including clustering, STS, bitext mining, and paraphrase detection.

\paragraph{Pruning and Knowledge Distillation} After stage 1 training, we prune the 0.6B model to three smaller sizes along three dimensions: hidden size, MLP intermediate size, and number of layers. For hidden size and MLP intermediate size, we prune the rows and columns in associated weight matrices based on activation norms on a small set of calibration data. For the layer dimension, we simply keep the first $n$ layers of the model. We also experimented with pruning layers based on the change of activation norms, but found it to underperform this simple method.

After pruning, we find that naive training leads to large performance drops (see Table~\ref{tab:results-distillation}). We mitigate this by applying an additional knowledge distillation loss when training the pruned models, computed by the MSE between the student's sequence embedding and a teacher's sequence embedding over input query, document, and negatives. Ablation experiments suggest that this form of knowledge distillation can also benefit larger models, so we apply it to the 0.6B and 1.7B models in the second training stage as well, while the three largest models are trained without distillation due to resource constraints.

All models are trained with AdamW optimizer~\citep{2019AdamW}. Matryoshka Representation Learning~\citep{2022MRL} is applied in both training stages, with a minimum matryoshka dimension of 8. The remaining training hyperparameters are given in Table~\ref{tab:model_config}
\section{Experiments}

\subsection{Main Results}

\begin{table*}[th]
    \centering
    \small
    \adjustbox{width=\textwidth,center}{
    \begin{tabular}{rccccccccc}
    \toprule
    \tb{Model} & \tb{Multi.}$^{\text{(131)}}$ & \tb{English}$^{\text{(41)}}$ & \tb{Code}$^{\text{(12)}}$ & \tb{Medical}$^{\text{(12)}}$ & \tb{European}$^{\text{(73)}}$ & \tb{Scan.}$^{\text{(28)}}$ & \tb{Indic}$^{\text{(20)}}$ & \tb{German}$^{\text{(19)}}$ & \tb{French}$^{\text{(25)}}$ \\
    \midrule
    \emph{14B} & 68.74 (6) & 73.08 (10) & 80.75 (1) & 65.20 (2) & 69.89 (1) & 71.10 (1) & 78.85 (1) & 67.02 (1) & 72.62 (2) \\
    \emph{8B} & 68.09 (8) & 72.86 (11) & 80.16 (5) & 64.91 (4) & 69.22 (2) & 69.94 (2) & 77.93 (2) & 66.81 (2) & 72.66 (1) \\
    \emph{4B} & 67.06 (10) & 72.41 (12) & 80.15 (6) & 64.48 (7) & 68.63 (3) & 68.46 (3) & 76.58 (3) & 66.06 (3) & 71.22 (3) \\
    \emph{1.7B} & 65.21 (13) & 71.63 (16) & 78.76 (8) & 61.40 (15) & 66.90 (4) & 67.21 (4) & 74.20 (4) & 65.08 (4) & 69.85 (5) \\
    \emph{0.6B} & 62.74 (17) & 69.97 (25) & 77.41 (10) & 57.95 (25) & 64.49 (5) & 64.32 (6) & 70.11 (6) & 63.08 (5) & 68.14 (7) \\
    \emph{330M} & 60.84 (26) & 68.86 (36) & 75.74 (13) & 56.44 (31) & 62.04 (13) & 61.93 (11) & 66.92 (11) & 61.61 (6) & 66.03 (13) \\
    \emph{160M} & 57.98 (38) & 65.93 (49) & 70.38 (18) & 52.39 (40) & 59.06 (22) & 57.79 (25) & 62.09 (20) & 57.35 (9) & 61.90 (20) \\
    \emph{80M} & 55.23 (50) & 64.55 (60) & 67.97 (22) & 50.74 (42) & 56.24 (35) & 55.54 (30) & 58.39 (34) & 55.56 (13) & 60.30 (22) \\
    \bottomrule
    \rowcolor{gray!30} \multicolumn{10}{c}{\textit{Results continued for remaining languages and average}} \\
    \toprule
    \tb{Model} & \tb{Korean}$^{\text{(6)}}$ & \tb{Polish}$^{\text{(17)}}$ & \tb{Chinese}$^{\text{(32)}}$ & \tb{Japan.}$^{\text{(28)}}$ & \tb{Dutch}$^{\text{(40)}}$ & \tb{Russian}$^{\text{(23)}}$ & \tb{Persian}$^{\text{(52)}}$ & \tb{Viet.}$^{\text{(50)}}$ & \tb{Avg.} \\
    \midrule
    \emph{14B} & 74.85 (3) & 75.13 (1) & 68.24 (21) & 79.32 (1) & 66.39 (1) & 70.90 (4) & 73.55 (1) & 63.56 (1) & 71.72 \\
    \emph{8B} & 75.11 (2) & 74.61 (2) & 67.73 (24) & 78.54 (2) & 65.81 (2) & 70.57 (5) & 72.69 (2) & 63.32 (2) & 71.23 \\
    \emph{4B} & 73.63 (5) & 73.42 (3) & 67.12 (27) & 77.43 (3) & 64.06 (5) & 69.46 (7) & 71.66 (3) & 62.74 (3) & 70.27 \\
    \emph{1.7B} & 73.77 (4) & 72.03 (4) & 66.41 (31) & 75.68 (4) & 62.73 (8) & 68.52 (9) & 70.01 (5) & 61.54 (4) & 68.88 \\
    \emph{0.6B} & 70.88 (7) & 69.63 (5) & 64.81 (35) & 73.07 (6) & 59.54 (10) & 65.97 (11) & 67.98 (7) & 59.56 (5) & 66.45 \\
    \emph{330M} & 68.70 (11) & 66.53 (6) & 63.02 (37) & 70.75 (9) & 57.57 (12) & 63.89 (15) & 66.14 (9) & 57.46 (6) & 64.38 \\
    \emph{160M} & 62.55 (18) & 62.32 (7) & 59.88 (40) & 65.74 (17) & 52.95 (26) & 59.71 (27) & 62.16 (17) & 52.56 (11) & 60.16 \\
    \emph{80M} & 59.98 (21) & 59.82 (8) & 58.22 (41) & 58.80 (19) & 50.54 (30) & 57.15 (35) & 59.98 (20) & 50.52 (15) & 57.62 \\
    \bottomrule
    \end{tabular}
    }
    \caption{Performance of F2LLM-v2 on 17 MTEB benchmarks and their rankings on the leaderboard, accessed on March 19th, 2026 and given in (parentheses). The number of tasks in each benchmark is given in $^{\text{(superscript)}}$.}
    \label{tab:main-results}
\end{table*}

We evaluate F2LLM-v2 on 17 MTEB benchmarks: Multilingual, English, Code, Medical, European, Scandinavian, Indic, German, French, Korean, Polish, Chinese, Japanese, Dutch, Russian, Persian, and Vietnamese, totaling 430 tasks across ten types: retrieval, reranking, classification, clustering, pair classification, multilabel classification, STS, instruction reranking, bitext mining, and summarization. More details on these benchmarks and tasks are given in Appendix~\ref{appendix:mteb}.

The main results are presented in Table~\ref{tab:main-results}, along with the models' rankings on the leaderboards. To assess the performance of the smaller models more thoroughly, we also compare specifically with individual models with the same sizes from the Qwen3-Embedding~\citep{2025Qwen3-Embedding} and EmbeddingGemma~\citep{2025EmbeddingGemma} families in Table~\ref{tab:mteb-comparison}. As the results of these models are not complete on several benchmarks, we use the results from the leaderboards when available, and evaluate them on the remaining tasks using the same prompts as those used to evaluate F2LLM-v2.

\begin{table*}[th]
    \centering
    \small
    \adjustbox{width=\textwidth,center}{
    \begin{tabular}{lccccccccc}
    \toprule
    \tb{Model} & \tb{Multi.}$^{\text{(131)}}$ & \tb{English}$^{\text{(41)}}$ & \tb{Code}$^{\text{(12)}}$ & \tb{Medical}$^{\text{(12)}}$ & \tb{European}$^{\text{(73)}}$ & \tb{Scan.}$^{\text{(28)}}$ & \tb{Indic}$^{\text{(20)}}$ & \tb{German}$^{\text{(19)}}$ & \tb{French}$^{\text{(25)}}$ \\
    \midrule
    \rowcolor{gray!15} \multicolumn{10}{c}{\textit{0.3B}}\\
    \emph{EmbedGemma} & \tb{61.15} & \tb{69.67} & 68.76 & 51.24 & \tb{62.50} & 54.39 & 66.11 & 56.28 & 61.90 \\
    \emph{F2LLM-v2} & 60.84 & 68.86 & \tb{75.74} & \tb{56.44} & 62.04 & \tb{61.93} & \tb{66.92} & \tb{61.61} & \tb{66.03} \\
    \midrule
    \rowcolor{gray!15} \multicolumn{10}{c}{\textit{0.6B}}\\
    \emph{Qwen3-Embed} & \tb{64.34} & \tb{70.47} & 75.42 & \tb{60.16} & 63.91 & 60.99 & 66.53 & 59.45 & 63.01 \\
    \emph{F2LLM-v2} & 62.74 & 69.97 & \tb{77.41} & 57.95 & \tb{64.49} & \tb{64.32} & \tb{70.11} & \tb{63.08} & \tb{68.14} \\
    \bottomrule
    \rowcolor{gray!30} \multicolumn{10}{c}{\textit{Results continued for remaining languages and average}} \\
    \toprule
    \tb{Model} & \tb{Korean}$^{\text{(6)}}$ & \tb{Polish}$^{\text{(17)}}$ & \tb{Chinese}$^{\text{(32)}}$ & \tb{Japan.}$^{\text{(28)}}$ & \tb{Dutch}$^{\text{(40)}}$ & \tb{Russian}$^{\text{(23)}}$ & \tb{Persian}$^{\text{(52)}}$ & \tb{Viet.}$^{\text{(50)}}$ & \tb{Avg.} \\
    \midrule
    \rowcolor{gray!15} \multicolumn{10}{c}{\textit{0.3B}}\\
    \emph{EmbedGemma} & 58.24 & 64.70 & 50.40 & 60.82 & 50.98 & \tb{64.57} & \tb{67.11} & 43.45 & 59.55 \\
    \emph{F2LLM-v2} & \tb{68.70} & \tb{66.53} & \tb{63.45} & \tb{70.75} & \tb{57.57} & 63.89 & 66.14 & \tb{57.46} & \tb{64.41} \\
    \midrule
    \rowcolor{gray!15} \multicolumn{10}{c}{\textit{0.6B}}\\
    \emph{Qwen3-Embed} & 65.29 & 67.42 & \tb{66.71} & 67.28 & 54.27 & 64.20 & 62.88 & 56.01 & 64.02 \\
    \emph{F2LLM-v2} & \tb{70.88} & \tb{69.63} & 65.23 & \tb{73.07} & \tb{59.54} & \tb{65.97} & \tb{67.98} & \tb{59.56} & \tb{66.47} \\
    \bottomrule
    \end{tabular}
    }
    \caption{Comparison of our models with EmbeddingGemma and Qwen3-Embedding. The number of tasks in each benchmark is given in $^{\text{(superscript)}}$.}
    \label{tab:mteb-comparison}
\end{table*}

These results highlight the scalability of the F2LLM-v2 models. Our 14B model achieves state-of-the-art on 11 of the 17 evaluated benchmarks, capturing deep semantic nuances suitable for enterprise-grade database systems where model inference does not present a bottleneck. In comparison, the smaller variants - particularly the 80M and 160M models - demonstrate remarkable efficiency, which is achieved without a proportional degradation in performance, verifying the effectiveness of our pruning and knowledge distillation pipeline. Notably, the 330M and 0.6B models consistently outperform Qwen3-Embedding and EmbeddingGemma on most language-specific benchmarks and the code benchmark, providing an ideal tradeoff between performance and efficiency for edge deployment.

\subsection{Ablation Studies}

For ablation studies, we conduct experiments on a subset of 350 tasks, which are selected based solely on evaluation time to speed up model iterations.

We first examine the effectiveness of knowledge distillation. Starting from the same stage-1 checkpoints, we train another series of models in identical settings as F2LLM-v2, but without knowledge distillation. The results are presented in Table~\ref{tab:results-distillation}, which demonstrates a consistent drop in performance across all five model scales, verifying the effectiveness of our knowledge distillation method in transferring the capabilities of teacher models into significantly more compact students.

\begin{table}[]
    \centering
    \small
    \begin{tabular}{cccccc}
    \toprule
        & \tb{80M} & \tb{160M} & \tb{330M} & \tb{0.6B} & \tb{1.7B} \\
    \midrule
        w. distillation (F2LLM-v2) & 58.04 & 60.53 & 64.55 & 66.72 & 69.13\\
        w.o. distillation & 53.37 & 56.27 & 62.77 & 65.87 & 68.58 \\
    \bottomrule
    \end{tabular}
    \caption{Ablation results on knowledge distillation, averaged over 350 tasks.}
    \label{tab:results-distillation}
\end{table}

\begin{figure}
    \centering
    \includegraphics[width=0.7\linewidth]{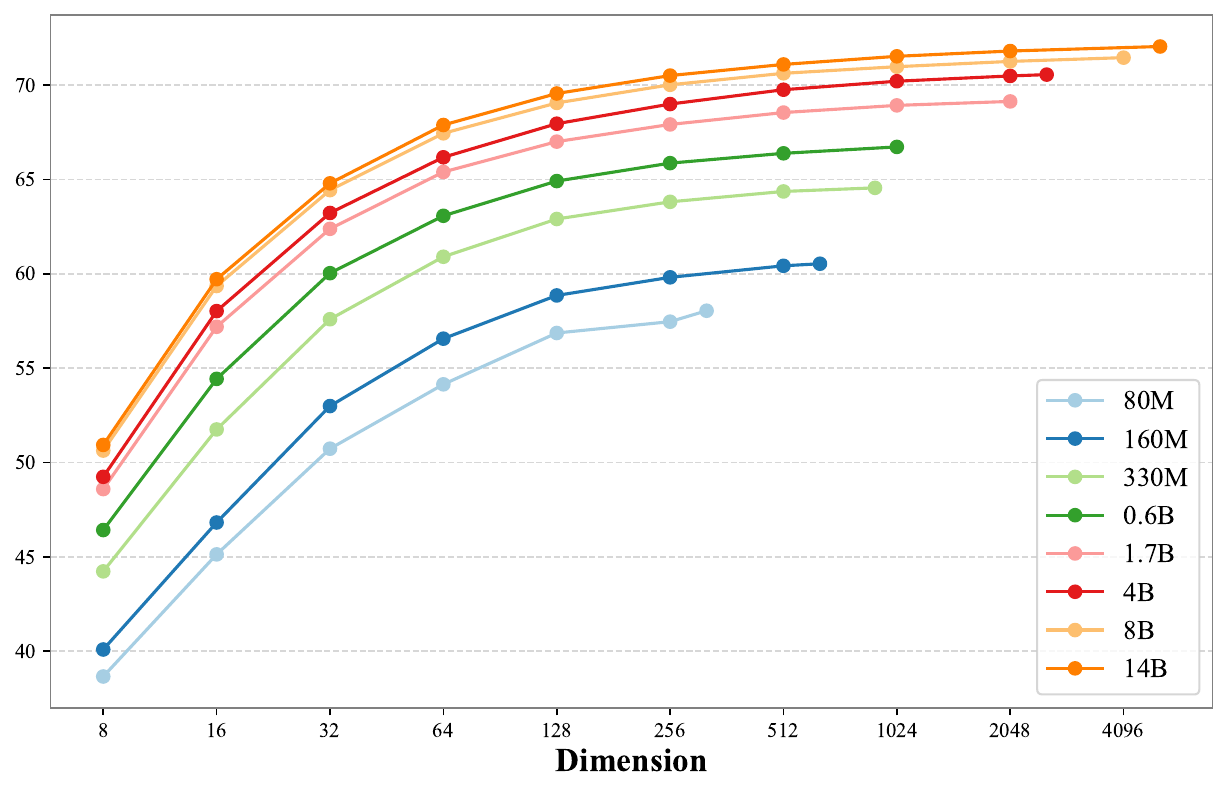}
    \caption{Results of evaluating F2LLM-v2 models at different representation sizes.}
    \label{fig:mrl_results}
\end{figure}


We also verify the effectiveness of MRL by evaluating the models at different representation sizes. For each model, we truncate the output embeddings to dimensions ranging from 8 to their full size and measure performance on the ablation task subset. The results, plotted in Figure~\ref{fig:mrl_results}, confirm that MRL successfully concentrates the most critical semantic information in the initial representation dimensions. Performance scales gracefully with the embedding dimension, with the steepest performance gains occurring at the lower dimensions from 8 to 128 and plateauing as the representation approaches its full size. This demonstrates that the leading dimensions effectively capture the most salient semantic features, while subsequent dimensions add progressively finer-grained detail.

These results highlight a crucial tradeoff for practitioners. For example, the 330M model using its full 896-dimensional embedding performs comparably to the much larger 8B and 14B models when their embeddings are truncated to 32 dimensions. This flexibility enables users to dynamically select an optimal balance between performance, inference cost, and storage cost, showcasing the practical utility of MRL for deploying high-quality embeddings across a wide spectrum of hardware constraints.
\section{Conclusion}

F2LLM-v2 is the latest member of the Codefuse embedding model family~\citep{2024D2LLM,2025f2llm,2025C2LLM}. By addressing the current gaps of language imbalance and training opacity in embedding model research, F2LLM-v2 represents a significant step forward in democratizing high-performance embedding models. With the release of 8 models along with the complete training recipe and intermediate checkpoints, we hope to facilitate transparency in frontier embedding research and contribute to a future with truly global equity in AI technology deployment.

\bibliographystyle{colm2024_conference}
\bibliography{custom}

\appendix

\clearpage
\section{Training Data Details}\label{appendix:data}

\begin{table}[th]
    \small
    \centering
    \adjustbox{width=\textwidth,center}{
    \begin{tabular}{m{1cm}<{\centering}cr|m{1cm}<{\centering}cr|m{1cm}<{\centering}cr}
    \toprule
        \tb{ISO Code} & \tb{Language} & \tb{Samples} & \tb{ISO Code} & \tb{Language} & \tb{Samples} & \tb{ISO Code} & \tb{Language} & \tb{Samples} \\
    \cmidrule(r){1-3}\cmidrule(lr){4-6}\cmidrule(l){7-9}
eng & English & 16,059,324 & msa & Malay & 175,111 & tat & Tatar & 22,327 \\
zho & Chinese & 4,280,372 & hrv & Croatian & 146,132 & bos & Bosnian & 21,175 \\
rus & Russian & 3,426,943 & isl & Icelandic & 124,495 & yor & Yoruba & 20,139 \\
spa & Spanish & 2,771,907 & slv & Slovenian & 122,065 & min & Minangkabau & 19,868 \\
fra & French & 2,426,075 & srp & Serbian & 113,663 & che & Chechen & 19,518 \\
deu & German & 1,749,922 & urd & Urdu & 113,258 & arz & Egyptian Arabic & 16,783 \\
ara & Arabic & 1,402,943 & ben & Bengali & 87,787 & lmo & Lombard & 16,575 \\
nld & Dutch & 1,395,135 & aze & Azerbaijani & 82,209 & arg & Aragonese & 16,500 \\
vie & Vietnamese & 1,159,472 & afr & Afrikaans & 81,233 & bak & Bashkir & 16,451 \\
hin & Hindi & 1,106,611 & tam & Tamil & 78,384 & som & Somali & 16,369 \\
kor & Korean & 1,083,205 & kat & Georgian & 77,567 & als & Tosk Albanian & 15,655 \\
jpn & Japanese & 1,082,466 & tel & Telugu & 77,362 & ido & Ido & 15,613 \\
ita & Italian & 960,595 & mal & Malayalam & 76,518 & szl & Silesian & 14,845 \\
ind & Indonesian & 952,218 & mon & Mongolian & 59,851 & wuu & Wu Chinese & 14,762 \\
por & Portuguese & 919,257 & nno & Norwegian Nynorsk & 58,298 & new & Nepal Bhasa & 14,714 \\
pol & Polish & 885,709 & kaz & Kazakh & 55,317 & chv & Chuvash & 13,759 \\
tur & Turkish & 694,051 & cym & Welsh & 53,951 & pnb & Western Panjabi & 13,657 \\
dan & Danish & 675,313 & mar & Marathi & 53,803 & fry & Western Frisian & 13,649 \\
tha & Thai & 654,534 & sqi & Albanian & 53,602 & snd & Sindhi & 13,210 \\
swe & Swedish & 577,848 & nob & Norwegian Bokmål & 52,905 & ori & Oriya & 12,791 \\
fas & Persian & 535,107 & pus & Pushto & 52,753 & plt & Plateau Malagasy & 12,717 \\
ukr & Ukrainian & 471,079 & mkd & Macedonian & 52,504 & scn & Sicilian & 12,694 \\
ces & Czech & 467,569 & hbs & Serbo-Croatian & 48,523 & kur & Kurdish & 11,872 \\
nor & Norwegian & 424,995 & ceb & Cebuano & 47,408 & sun & Sundanese & 11,793 \\
ell & Modern Greek & 378,202 & jav & Javanese & 47,283 & bar & Bavarian & 11,193 \\
cat & Catalan & 370,156 & war & Waray (Philippines) & 45,348 & yid & Yiddish & 10,785 \\
ron & Romanian & 344,845 & kan & Kannada & 44,534 & ckb & Central Kurdish & 9,829 \\
fin & Finnish & 332,394 & epo & Esperanto & 44,266 & fao & Faroese & 9,825 \\
bul & Bulgarian & 321,379 & lat & Latin & 43,335 & ina & Interlingua & 9,782 \\
tgl & Tagalog & 313,985 & guj & Gujarati & 40,184 & gla & Scottish Gaelic & 9,769 \\
glg & Galician & 301,386 & uzb & Uzbek & 39,951 & bug & Buginese & 9,662 \\
mya & Burmese & 294,167 & amh & Amharic & 38,763 & que & Quechua & 9,406 \\
hye & Armenian & 288,622 & oci & Occitan & 37,413 & bpy & Bishnupriya & 9,400 \\
khm & Khmer & 287,530 & bel & Belarusian & 33,330 & san & Sanskrit & 8,730 \\
nep & Nepali & 276,057 & azb & South Azerbaijani & 31,815 & lim & Limburgan & 8,573 \\
hun & Hungarian & 271,802 & kir & Kirghiz & 29,319 & hau & Hausa & 8,435 \\
eus & Basque & 270,551 & mlg & Malagasy & 28,661 & mai & Maithili & 8,180 \\
heb & Hebrew & 263,869 & vol & Volapük & 27,187 & zsm & Standard Malay & 8,179 \\
lao & Lao & 244,750 & ast & Asturian & 26,004 & ibo & Igbo & 8,132 \\
swa & Swahili & 241,497 & pan & Panjabi & 25,096 & vec & Venetian & 8,121 \\
azj & North Azerbaijani & 213,046 & ltz & Luxembourgish & 25,092 & ilo & Iloko & 7,968 \\
lav & Latvian & 212,058 & nds & Low German & 24,713 & asm & Assamese & 7,042 \\
sin & Sinhala & 207,903 & hat & Haitian & 23,940 & sah & Yakut & 7,011 \\
slk & Slovak & 202,049 & bre & Breton & 23,931 & arb & Standard Arabic & 6,945 \\
tgk & Tajik & 200,631 & gle & Irish & 23,148 & sna & Shona & 6,933 \\
est & Estonian & 191,063 & sco & Scots & 23,032 & mlt & Maltese & 6,911 \\
lit & Lithuanian & 184,391 & xho & Xhosa & 22,799 & zul & Zulu & 6,669 \\
    \bottomrule
    \end{tabular}
    }
    \caption{Natural language distribution in the training data of F2LLM-v2 (part1).}
    \label{tab:language-distribution-1}
\end{table}

\begin{table}[th]
    \small
    \centering
    \adjustbox{width=\textwidth,center}{
    \begin{tabular}{m{1cm}<{\centering}cr|m{1cm}<{\centering}cr|m{1cm}<{\centering}cr}
    \toprule
        \tb{ISO Code} & \tb{Language} & \tb{Samples} & \tb{ISO Code} & \tb{Language} & \tb{Samples} & \tb{ISO Code} & \tb{Language} & \tb{Samples} \\
    \cmidrule(r){1-3}\cmidrule(lr){4-6}\cmidrule(l){7-9}
mzn & Mazanderani & 6,352 & tsn & Tswana & 1,539 & lvs & Standard Latvian & 800 \\
uig & Uighur & 6,190 & mwl & Mirandese & 1,491 & mag & Magahi & 800 \\
oss & Iron Ossetic & 5,893 & div & Dhivehi & 1,387 & mni & Manipuri & 800 \\
tuk & Turkmen & 5,854 & kbp & Kabiyè & 1,349 & mos & Mossi & 800 \\
ary & Moroccan Arabic & 5,703 & chm & Mari & 1,238 & nqo & N'Ko & 800 \\
wln & Walloon & 5,408 & ewe & Ewe & 1,220 & nus & Nuer & 800 \\
cdo & Min Dong Chinese & 5,175 & smo & Samoan & 1,175 & ory & Odia & 800 \\
npi & Nepali & 5,156 & tso & Tsonga & 1,174 & prs & Dari & 800 \\
nap & Neapolitan & 4,778 & fij & Fijian & 1,122 & quy & Ayacucho Quechua & 800 \\
ace & Achinese & 4,758 & bam & Bambara & 1,061 & sat & Santali & 800 \\
mrj & Western Mari & 4,728 & lin & Lingala & 1,046 & tpi & Tok Pisin & 800 \\
xmf & Mingrelian & 4,712 & nav & Navajo & 1,028 & tum & Tumbuka & 800 \\
pes & Iranian Persian & 4,414 & roh & Romansh & 999 & tzm & Central Atlas Tamazight & 800 \\
diq & Dimli & 4,140 & ssw & Swati & 982 & umb & Umbundu & 800 \\
apc & Levantine Arabic & 4,079 & awa & Awadhi & 954 & uzn & Northern Uzbek & 800 \\
wol & Wolof & 4,068 & pag & Pangasinan & 954 & ydd & Eastern Yiddish & 800 \\
pbt & Southern Pashto & 3,796 & cor & Cornish & 938 & yue & Yue Chinese & 800 \\
nso & Pedi & 3,676 & dzo & Dzongkha & 928 & dyu & Dyula & 799 \\
srd & Sardinian & 3,614 & udm & Udmurt & 890 & lua & Luba-Lulua & 799 \\
ban & Balinese & 3,581 & fon & Fon & 883 & twi & Twi & 798 \\
lij & Ligurian & 3,487 & kon & Kongo & 873 & aeb & Tunisian Arabic & 793 \\
hsb & Upper Sorbian & 3,440 & glv & Manx & 864 & kik & Kikuyu & 761 \\
acq & Ta'izzi-Adeni Arabic & 3,188 & tir & Tigrinya & 851 & tyv & Tuvinian & 626 \\
crh & Crimean Tatar & 2,985 & pms & Piemontese & 842 & ava & Avaric & 588 \\
mri & Maori & 2,922 & myv & Erzya & 840 & aym & Aymara & 587 \\
egl & Emilian & 2,859 & sag & Sango & 827 & krc & Karachay-Balkar & 587 \\
ars & Najdi Arabic & 2,712 & run & Rundi & 825 & ful & Fulah & 581 \\
grn & Guarani & 2,705 & acm & Mesopotamian Arabic & 800 & orm & Oromo & 548 \\
nya & Chichewa & 2,607 & aka & Akan & 800 & stq & Saterfriesisch & 461 \\
hif & Fiji Hindi & 2,566 & ayr & Central Aymara & 800 & lah & Lahnda & 450 \\
kas & Kashmiri & 2,363 & bem & Bemba & 800 & ton & Tonga & 391 \\
fur & Friulian & 2,284 & bho & Bhojpuri & 800 & mdf & Moksha & 314 \\
swh & Swahili & 2,253 & cjk & Chokwe & 800 & haw & Hawaiian & 299 \\
fil & Filipino & 2,156 & dik & Southwestern Dinka & 800 & nia & Nias & 297 \\
sme & Northern Sami & 2,156 & fuv & Nigerian Fulfulde & 800 & bis & Bislama & 272 \\
shn & Shan & 2,098 & gaz & West Central Oromo & 800 & alt & Southern Altai & 250 \\
sot & Southern Sotho & 2,089 & hne & Chhattisgarhi & 800 & srn & Sranan Tongo & 204 \\
kin & Kinyarwanda & 2,031 & kab & Kabyle & 800 & ven & Venda & 194 \\
lug & Ganda & 1,994 & kac & Kachin & 800 & kbd & Kabardian & 172 \\
pap & Papiamento & 1,974 & kam & Kamba (Kenya) & 800 & xal & Kalmyk & 122 \\
cos & Corsican & 1,949 & kea & Kabuverdianu & 800 & din & Dinka & 104 \\
mhr & Eastern Mari & 1,633 & khk & Halh Mongolian & 800 & jam & Jamaican Creole English & 100 \\
bjn & Banjar & 1,600 & kmb & Kimbundu & 800 & kal & Kalaallisut & 92 \\
knc & Central Kanuri & 1,600 & kmr & Northern Kurdish & 800 & iku & Inuktitut & 84 \\
taq & Tamasheq & 1,600 & ltg & Latgalian & 800 & guc & Wayuu & 52 \\
kom & Komi & 1,583 & luo & Luo & 800 & chr & Cherokee & 51 \\
bod & Tibetan & 1,563 & lus & Lushai & 800 & ady & Adyghe & 33 \\
    \bottomrule
    \end{tabular}
    }
    \caption{Natural language distribution in the training data of F2LLM-v2 (part2).}
    \label{tab:language-distribution-2}
\end{table}

\begin{table}[th]
    \small
    \centering
    \begin{tabular}{cr|cr}
    \toprule
Language & Samples & Language & Samples \\
    \cmidrule(r){1-2}\cmidrule(lr){3-4}
python & 1,972,390 & css & 1,003 \\
php & 553,651 & typescript & 888 \\
java & 483,469 & r & 636 \\
cpp & 393,514 & lisp & 467 \\
go & 351,586 & jsx & 436 \\
javascript & 245,632 & objective-c & 327 \\
c\# & 92,008 & json & 264 \\
ruby & 68,317 & xml & 258 \\
c & 43,487 & yaml & 180 \\
rust & 12,924 & assembly & 171 \\
kotlin & 11,284 & powershell & 162 \\
sql & 6,826 & vba & 157 \\
pascal & 6,299 & lua & 126 \\
d & 5,278 & matlab & 114 \\
haskell & 4,967 & dart & 107 \\
scala & 4,120 & bash & 105 \\
html & 2,777 & http & 99 \\
shell & 2,095 & graphql & 89 \\
perl & 2,009 & svg & 82 \\
swift & 1,907 & vb.net & 75 \\
ocaml & 1,894 & groovy & 63 \\
csharp & 1,891 & Misc. & 2,301 \\
    \bottomrule
    \end{tabular}
    \caption{Programming language distribution in the training data of F2LLM-v2.}
    \label{tab:language-distribution-3}
\end{table}
\begin{table}[th]
    \tiny
    \centering
    \adjustbox{width=\textwidth,center}{
    \begin{tabular}{lccrm{9cm}}
    \toprule
        \tb{Name} & \tb{Language} & \tb{Format} & \tb{Size} & \tb{URL} \\
    \midrule
\rowcolor{gray!15} \multicolumn{5}{c}{Bitext Mining}\\
UNPC~\citep{2016unpc} & 6 & Retrieval & 2,922,245 & \url{huggingface.co/datasets/Helsinki-NLP/un_pc} \\
ParaCrawl~\citep{2020paracrawl} & 30 & Retrieval & 10,684,184 & \url{paracrawl.eu/index.php} \\
BactrianX Translation~\citep{2023bactrianx} & 52 & Clustering & 491,282 & \url{huggingface.co/datasets/MBZUAI/Bactrian-X} \\
Europarl~\citep{2005europarl} & 21 & Clustering & 477,566 & \url{huggingface.co/datasets/Helsinki-NLP/europarl} \\
\midrule
\rowcolor{gray!15} \multicolumn{5}{c}{Question Answering}\\
WebFAQ~\citep{2025webfaq} & 49 & Retrieval & 4,368,504 & \url{huggingface.co/datasets/PaDaS-Lab/webfaq-retrieval} \\
mMARCO~\citep{2021mmarco} & 14 & Retrieval & 5,470,174 & \url{huggingface.co/datasets/unicamp-dl/mmarco} \\
PAQ~\citep{2021PAQ} & en & Retrieval & 938,771 & \url{huggingface.co/datasets/sentence-transformers/paq} \\
SQuAD~\citep{2016SQuAD} & en & Retrieval & 89,509 & \url{huggingface.co/datasets/rajpurkar/squad} \\
Stack Exchange~\citep{2021StackExchangeDataset} & en & Retrieval & 754,705 & \url{huggingface.co/datasets/flax-sentence-embeddings/stackexchange_titlebody_best_voted_answer_jsonl} \\
Arguana~\citep{2018Arguana} & en & Retrieval & 22,848 & \url{huggingface.co/datasets/BeIR/arguana-generated-queries} \\
Natural Questions~\citep{2019NaturalQuestions} & en & Retrieval & 97,209 & \url{huggingface.co/datasets/sentence-transformers/natural-questions} \\
HotpotQA~\citep{2018HotpotQA} & en & Retrieval & 120,528 & \url{huggingface.co/datasets/mteb/hotpotqa} \\
ELI5~\citep{2019ELI5} & en & Retrieval & 161,345 & \url{huggingface.co/datasets/Pavithree/eli5} \\
FiQA2018~\citep{2018FIQA} & en & Retrieval & 7,452 & \url{huggingface.co/datasets/mteb/fiqa} \\
BioASQ~\citep{2015BioASQ} & en & Retrieval & 125,248 & \url{huggingface.co/datasets/BeIR/bioasq-generated-queries} \\
NFCorpus~\citep{2016NFCorpus} & en & Retrieval & 1,283 & \url{huggingface.co/datasets/mteb/nfcorpus} \\
TriviaQA~\citep{2017TriviaQA} & en & Retrieval & 60,025 & \url{huggingface.co/datasets/sentence-transformers/trivia-qa-triplet} \\
PubMedQA~\citep{2019PubMedQA} & en & Retrieval & 60,227 & \url{huggingface.co/datasets/qiaojin/PubMedQA} \\
Amazon QA~\citep{2019AmazonQA} & en & Retrieval & 59,340 & \url{github.com/amazonqa/amazonqa} \\
MIRACL~\citep{2023MIRACL} & 16 & Retrieval & 26,740 & \url{huggingface.co/datasets/miracl/miracl} \\
Mr.TyDi~\citep{2021mrtidy} & 11 & Retrieval & 48,619 & \url{huggingface.co/datasets/mteb/mrtidy} \\
MLDR~\citep{2024BGE-M3} & 13 & Retrieval & 40,264 & \url{huggingface.co/datasets/Shitao/MLDR} \\
MKQA~\citep{2021mkqa} & 26 & Retrieval & 69,287 & \url{huggingface.co/datasets/mteb/MKQARetrieval} \\
StackOverflowQA~\citep{2025CoIR} & en & Retrieval & 13,820 & \url{huggingface.co/datasets/mteb/stackoverflow-qa} \\
ProCQA~\citep{2025procqa} & 11 & Retrieval & 485,780 & \url{github.com/jordane95/procqa} \\
Yahoo\_Answers~\citep{2015yahooanswers} & en & Retrieval & 196,645 & \url{huggingface.co/datasets/sentence-transformers/yahoo-answers} \\
GooAQ~\citep{2021gooaq} & en & Retrieval & 473,876 & \url{github.com/allenai/gooaq} \\
T2Ranking~\citep{2023t2ranking} & zh & Retrieval & 85,521 & \url{huggingface.co/datasets/sentence-transformers/t2ranking} \\
DuReader~\citep{2018dureader} & zh & Retrieval & 78,023 & \url{huggingface.co/datasets/sentence-transformers/dureader} \\
cMedQAv2~\citep{2018cmedqav2} & zh & Retrieval & 23,105 & \url{huggingface.co/datasets/sentence-transformers/cmedqa-v2} \\
Huatuo\_kgqa~\citep{2025huatuoqa} & zh & Retrieval & 53,835 & \url{huggingface.co/datasets/FreedomIntelligence/huatuo_knowledge_graph_qa} \\
Huatuo\_encqa~\citep{2025huatuoqa} & zh & Retrieval & 253,523 & \url{huggingface.co/datasets/FreedomIntelligence/huatuo_encyclopedia_qa} \\
Multi CPR Medical~\citep{2022multicprmedical} & zh & Retrieval & 62,085 & \url{github.com/Alibaba-NLP/Multi-CPR} \\
HealthCareMagic~\citep{2023healthcaremagic} & en & Retrieval & 78,626 & \url{github.com/Kent0n-Li/ChatDoctor} \\
MedicalQA\_ru~\citep{2022medicalqaru} & ru & Retrieval & 71,932 & \url{huggingface.co/datasets/blinoff/medical_qa_ru_data} \\
LLM Retrieval Data~\citep{2024llm_retrieval_data} & zh & Retrieval & 177,850 & \url{huggingface.co/datasets/infgrad/retrieval_data_llm} \\
RefGPT~\citep{2023refgpt} & zh & Retrieval & 184,332 & \url{github.com/sufengniu/RefGPT} \\
Lawzhidao~\citep{2020lawzhidao} & zh & Retrieval & 11,899 & \url{heywhale.com/mw/dataset/5e953ca8e7ec38002d02fca7/content} \\
MedMCQA~\citep{2022MedMCQA} & en & Retrieval & 16,526 & \url{huggingface.co/datasets/openlifescienceai/medmcqa} \\
CMCQA~\citep{2022CMCQA} & zh & Retrieval & 108,529 & \url{github.com/WENGSYX/CMCQA} \\
MedQA~\citep{2020MedQA} & en, zh & Retrieval & 13,458 & \url{github.com/jind11/MedQA} \\
webMedQA~\citep{2019webMedQA} & zh & Retrieval & 27,122 & \url{github.com/hejunqing/webMedQA} \\
MedQuAD~\citep{2019MedQuAD} & en & Retrieval & 14,268 & \url{github.com/abachaa/MedQuAD} \\
Medical Flashcards~\citep{2023medical_flashcards} & en & Retrieval & 33,183 & \url{huggingface.co/datasets/medalpaca/medical_meadow_medical_flashcards} \\
MailruQA~\citep{2022mailruqa} & ru & Retrieval & 150,777 & \url{huggingface.co/datasets/Den4ikAI/mailruQA-big} \\
WikiOmnia~\citep{2022wikiomnia} & ru & Retrieval & 462,693 & \url{huggingface.co/datasets/RussianNLP/wikiomnia} \\
PersianQA~\citep{2021PersianQA} & fa & Retrieval & 6,277 & \url{github.com/SajjjadAyobi/PersianQA} \\
PQuAD~\citep{2023PQuAD} & fa & Retrieval & 46,699 & \url{github.com/AUT-NLP/PQuAD} \\
ParSQuAD~\citep{2021parsquad} & fa & Retrieval & 41,879 & \url{github.com/BigData-IsfahanUni/ParSQuAD} \\
MQA~\citep{2021mqa} & 38 & Retrieval & 5,131,895 & \url{huggingface.co/datasets/clips/mqa} \\
HotpotQA-NL~\citep{2025beir-nl} & nl & Retrieval & 81,192 & \url{huggingface.co/datasets/clips/beir-nl-hotpotqa} \\
\midrule
\rowcolor{gray!15} \multicolumn{5}{c}{Instruction Data}\\
Aya~\citep{2024aya} & 65 & Retrieval & 126,965 & \url{huggingface.co/datasets/CohereLabs/aya_dataset} \\
MURI~\citep{2025muri} & 194 & Retrieval & 720,782 & \url{huggingface.co/datasets/akoksal/muri-it} \\
OASST2~\citep{2023oasst2} & 26 & Retrieval & 12,449 & \url{huggingface.co/datasets/OpenAssistant/oasst2} \\
MultiAlpaca~\citep{2023multialpaca} & 11 & Retrieval & 125,447 & \url{huggingface.co/datasets/DAMO-NLP-MT/multialpaca} \\
WildChat~\cite{2024wildchat} & 76 & Retrieval & 638,781 & \url{huggingface.co/datasets/allenai/WildChat-4.8M} \\
M2Lingual~\citep{2025m2lingual} & 75 & Retrieval & 158,251 & \url{huggingface.co/datasets/ServiceNow-AI/M2Lingual} \\
Natural Reasoning~\citep{2025naturalreasoning} & en & Retrieval & 845,682 & \url{huggingface.co/datasets/facebook/natural_reasoning} \\
Infinity Instruct~\citep{2025infinityinstruct} & en, zh & Retrieval & 757,439 & \url{huggingface.co/datasets/BAAI/Infinity-Instruct} \\
COIG~\citep{2025coig} & zh & Retrieval & 42,415 & \url{huggingface.co/datasets/m-a-p/COIG-CQIA} \\
Medinstruct~\citep{2023medinstruct} & en & Retrieval & 51,539 & \url{github.com/XZhang97666/AlpaCare} \\
CodeFeedbackST~\citep{2025CoIR} & 137 & Retrieval & 115,971 & \url{huggingface.co/datasets/mteb/codefeedback-st} \\
CodeFeedbackMT~\citep{2025CoIR} & python & Retrieval & 52,221 & \url{huggingface.co/datasets/mteb/codefeedback-mt} \\
OpenOrca~\citep{2023openorca} & en & Retrieval & 896,450 & \url{huggingface.co/datasets/Open-Orca/OpenOrca} \\
MEDI2~\citep{2025medi2} & en & Retrieval & 668,036 & \url{huggingface.co/datasets/GritLM/MEDI2} \\
MedicalInstruction~\citep{2023MedicalInstruction} & en & Retrieval & 75,268 & \url{huggingface.co/datasets/Mohammed-Altaf/medical-instruction-120k} \\
SiberianDataset~\citep{2023SiberianDatasetXL} & ru & Retrieval & 255,663 & \url{huggingface.co/datasets/SiberiaSoft/SiberianDatasetXL} \\
Ru Instruct~\citep{2024ru-instruct-translated} & ru & Retrieval & 452,574 & \url{huggingface.co/datasets/d0rj/ru-instruct} \\
KoAlpaca-RealQA~\citep{2024koalpaca_realqa} & ko & Retrieval & 17,599 & \url{huggingface.co/datasets/beomi/KoAlpaca-RealQA} \\
KoAlpaca~\citep{2023koalpaca} & ko & Retrieval & 21,126 & \url{huggingface.co/datasets/beomi/KoAlpaca-v1.1a} \\
KoMagpie~\citep{2024komagpie} & ko & Retrieval & 428,780 & \url{huggingface.co/datasets/channelcorp/KoMagpie-raw} \\
Nordic Text Matching~\citep{2025nordic_text_matching} & da, sv, no & Retrieval & 182,485 & \url{huggingface.co/datasets/kardosdrur/synthetic-nordic-text_matching} \\
Nordic Retrieval~\citep{2025nordic_retrieval} & da, sv, no & Retrieval & 172,437 & \url{huggingface.co/datasets/kardosdrur/synthetic-nordic-retrieval} \\
Nordic Classification~\citep{2025nordic_classification} & da, sv, no & Classification & 199,280 & \url{huggingface.co/datasets/kardosdrur/synthetic-nordic-classification} \\
\midrule
\rowcolor{gray!15} \multicolumn{5}{c}{Title Matching}\\
S2ORC-Title-Abstract~\citep{2020S2ORC} & en & Retrieval & 250,000 & \url{huggingface.co/datasets/sentence-transformers/s2orc} \\
CORD 19~\citep{2020cord19} & en & Retrieval & 373,674 & \url{huggingface.co/datasets/medalpaca/medical_meadow_cord19} \\
Multi CPR ECom~\citep{2022multicprmedical} & zh & Retrieval & 90,850 & \url{github.com/Alibaba-NLP/Multi-CPR} \\
ESCI~\citep{2022esci} & en, ja, es & Retrieval & 80,468 & \url{huggingface.co/datasets/tasksource/esci} \\
CLIRMatrix~\citep{2020clirmatrix} & 137 & Retrieval & 3,275,561 & \url{github.com/ssun32/CLIRMatrix} \\
DBPedia~\citep{2017dbpedia} & en & Retrieval & 288,736 & \url{huggingface.co/datasets/BeIR/dbpedia-entity-generated-queries} \\
\midrule
\rowcolor{gray!15} \multicolumn{5}{c}{NLI}\\
SNLI~\citep{2015SNLI} & en & Retrieval & 54,585 & \url{huggingface.co/datasets/stanfordnlp/snli} \\
MNLI~\citep{2018MNLI} & en & Retrieval & 112,075 & \url{huggingface.co/datasets/nyu-mll/multi_nli} \\
ANLI~\citep{2020ANLI} & en & Retrieval & 18,801 & \url{huggingface.co/datasets/facebook/anli} \\
XNLI~\citep{2018XNLI} & 14 & Retrieval & 1,400,600 & \url{huggingface.co/datasets/mteb/xnli} \\
OCNLI~\citep{2020OCNLI} & zh & Retrieval & 6,616 & \url{huggingface.co/datasets/dirtycomputer/OCNLI} \\
CMNLI~\citep{2024cmnli} & zh & Retrieval & 113,914 & \url{huggingface.co/datasets/fenffef/cmnli} \\
MSciNLI~\citep{2024MSciNLI} & en & Retrieval & 19,185 & \url{huggingface.co/datasets/sadat2307/MSciNLI} \\
    \bottomrule
    \end{tabular}
    }
    \caption{Number of samples in our collected training dataset (part 1).}
    \label{tab:data-1}
\end{table}

\begin{table}[th]
    \tiny
    \centering
    \adjustbox{width=\textwidth,center}{
    \begin{tabular}{lccrm{9cm}}
    \toprule
        \tb{Name} & \tb{Language} & \tb{Format} & \tb{Size} & \tb{URL} \\
    \midrule
\rowcolor{gray!15} \multicolumn{5}{c}{Topic Classification}\\
Arxiv Clustering P2P~\citep{2022mteb-arxiv} & en & Clustering & 83,476 & \url{huggingface.co/datasets/mteb/raw_arxiv} \\
Arxiv Clustering S2S~\citep{2022mteb-arxiv} & en & Clustering & 83,486 & \url{huggingface.co/datasets/mteb/raw_arxiv} \\
Biorxiv Clustering P2P~\citep{2022mteb-biorxiv} & en & Clustering & 57,296 & \url{huggingface.co/datasets/mteb/raw_biorxiv} \\
Biorxiv Clustering S2S~\citep{2022mteb-biorxiv} & en & Clustering & 57,296 & \url{huggingface.co/datasets/mteb/raw_biorxiv} \\
Medrxiv Clustering P2P~\citep{2022mteb-medrxiv} & en & Clustering & 18,659 & \url{huggingface.co/datasets/mteb/raw_medrxiv} \\
Medrxiv Clustering S2S~\citep{2022mteb-medrxiv} & en & Clustering & 18,659 & \url{huggingface.co/datasets/mteb/raw_medrxiv} \\
MLSUM Clustering~\citep{2020mlsum} & de, es, fr, ru & Clustering & 325,739 & \url{huggingface.co/datasets/mteb/mlsum} \\
TwentyNewsgroups~\citep{1995TwentyNewsGroups} & en & Clustering & 11,060 & \url{huggingface.co/datasets/SetFit/20_newsgroups} \\
SIB200ClusteringS2S~\citep{2024SIB200} & 205 & Clustering & 163,302 & \url{huggingface.co/datasets/mteb/sib200} \\
Reddit Clustering P2P~\citep{2021Reddit-Clustering-P2P} & en & Clustering & 80,000 & \url{huggingface.co/datasets/sentence-transformers/reddit-title-body} \\
Reddit Clustering S2S~\citep{2021Reddit-StackExchange-Clustering-S2S} & en & Clustering & 58,141 & \url{github.com/UKPLab/TWEAC-qa-agent-selection/tree/master/data/reddit/train} \\
Stack Exchange Clustering P2P~\citep{2021StackExchange-Clustering-P2P} & en & Clustering & 80,000 & \url{huggingface.co/datasets/flax-sentence-embeddings/stackexchange_title_body_jsonl} \\
Stack Exchange Clustering S2S~\citep{2021Reddit-StackExchange-Clustering-S2S} & en & Clustering & 56,731 & \url{github.com/UKPLab/TWEAC-qa-agent-selection/tree/master/data/stackexchange/train} \\
THUCNews~\citep{2016THUCNews} & zh & Clustering & 100,000 & \url{huggingface.co/datasets/SirlyDreamer/THUCNews} \\
TNews~\citep{2020CLUE} & zh & Clustering & 49,726 & \url{huggingface.co/datasets/C-MTEB/TNews-classification} \\
CSL~\citep{2022CSL} & zh & Clustering & 100,000 & \url{huggingface.co/datasets/neuclir/csl} \\
\midrule
\rowcolor{gray!15} \multicolumn{5}{c}{Summarization}\\
XSum~\citep{2018XSum} & en & Retrieval & 184,383 & \url{huggingface.co/datasets/EdinburghNLP/xsum} \\
CNN\_DM~\citep{2015CNN-DM} & en & Retrieval & 100,000 & \url{huggingface.co/datasets/abisee/cnn_dailymail} \\
MLSUM Retreival~\citep{2020mlsum} & 5 & Retrieval & 801,159 & \url{huggingface.co/datasets/mteb/mlsum} \\
Sentence Compression~\citep{2013Sentence-Compression} & en & Retrieval & 175,477 & \url{huggingface.co/datasets/sentence-transformers/sentence-compression} \\
\midrule
\rowcolor{gray!15} \multicolumn{5}{c}{Text-to-Code}\\
OCGI~\citep{2024OCGI} & python & Retrieval & 1,052,849 & \url{huggingface.co/datasets/nvidia/OpenCodeGeneticInstruct} \\
OpenCodeReasoning-2~\citep{2025OpenCodeReasoning2} & python, cpp & Retrieval & 16,632 & \url{huggingface.co/datasets/nvidia/OpenCodeReasoning-2} \\
xCodeEval NL2Code~\citep{2024xcodeeval} & 17 & Retrieval & 51,072 & \url{huggingface.co/datasets/NTU-NLP-sg/xCodeEval} \\
CosQA~\citep{2021CosQA} & python & Retrieval & 9,409 & \url{huggingface.co/datasets/mteb/cosqa} \\
SyntheticText2SQL~\citep{2024synthetic-text-to-sql} & sql & Retrieval & 99,617 & \url{huggingface.co/datasets/mteb/synthetic-text2sql} \\
\midrule
\rowcolor{gray!15} \multicolumn{5}{c}{Code-to-Code}\\
xCodeEval Code2Code~\citep{2024xcodeeval} & 17 & Retrieval & 37,056 & \url{huggingface.co/datasets/NTU-NLP-sg/xCodeEval} \\
xCodeEval Translation~\citep{2024xcodeeval} & 11 & Clustering & 500,000 & \url{huggingface.co/datasets/NTU-NLP-sg/xCodeEval} \\
CodeSearchNet-ccr~\citep{2025CoIR} & 6 & Retrieval & 905,195 & \url{huggingface.co/datasets/CoIR-Retrieval/CodeSearchNet-ccr} \\
\midrule
\rowcolor{gray!15} \multicolumn{5}{c}{Code-to-Text}\\
CodeSearchNet~\citep{2019CodeSearchNet} & 6 & Retrieval & 936,813 & \url{huggingface.co/datasets/CoIR-Retrieval/CodeSearchNet} \\
\midrule
\rowcolor{gray!15} \multicolumn{5}{c}{Paraphrase Detection}\\
StackExchangeDupQuestions-S2S~\citep{2021Embedding-Training-Data} & en & Retrieval & 183,559 & \url{huggingface.co/datasets/sentence-transformers/stackexchange-duplicates} \\
StackExchangeDupQuestions-P2P~\citep{2021Embedding-Training-Data} & en & Retrieval & 203,060 & \url{huggingface.co/datasets/sentence-transformers/stackexchange-duplicates} \\
QQP~\citep{2019QQP} & en & Retrieval & 243,598 & \url{gluebenchmark.com/tasks} \\
StackOverflowDupQuestions~\citep{2018StackOverflowDupQuestions} & en & Retrieval & 19,847 & \url{huggingface.co/datasets/mteb/stackoverflowdupquestions-reranking} \\
PawsX~\citep{2019pawsx} & 7 & Retrieval & 216,219 & \url{huggingface.co/datasets/google-research-datasets/paws-x} \\
LCQMC~\citep{2018LCQMC} & zh & Retrieval & 167,213 & \url{huggingface.co/datasets/C-MTEB/LCQMC} \\
\midrule
\rowcolor{gray!15} \multicolumn{5}{c}{Sentiment Analysis}\\
Amazon Polarity~\citep{2013Amazon-Reviews} & en & Classification & 100,000 & \url{huggingface.co/datasets/mteb/amazon_polarity} \\
IMDb~\citep{2011IMDB} & en & Classification & 24,904 & \url{huggingface.co/datasets/mteb/imdb} \\
Toxic Conversations~\citep{2019Toxic-Conversations} & en & Classification & 49,900 & \url{huggingface.co/datasets/mteb/toxic_conversations_50k} \\
Amazon Counterfactual~\citep{2021Amazon-Counterfactual} & en, de, ja & Classification & 14,870 & \url{huggingface.co/datasets/mteb/amazon_counterfactual} \\
Waimai~\citep{2023CMTEB} & zh & Classification & 7,999 & \url{huggingface.co/datasets/C-MTEB/waimai-classification} \\
Amazon Reviews~\citep{2013Amazon-Reviews} & 6 & Clustering & 600,000 & \url{huggingface.co/datasets/mteb/amazon_reviews_multi} \\
Emotion~\citep{2018Emotion} & en & Clustering & 17,944 & \url{huggingface.co/datasets/mteb/emotion} \\
Tweet Sentiment Extraction~\citep{2020tweet-sentiment-extraction} & en & Clustering & 26,732 & \url{huggingface.co/datasets/mteb/tweet_sentiment_extraction} \\
RuSentiment~\citep{2021ru_sentiment} & ru & Clustering & 100,000 & \url{huggingface.co/datasets/MonoHime/ru_sentiment_dataset} \\
CEDR~\citep{2020cedr} & ru & Clustering & 4,376 & \url{huggingface.co/datasets/sagteam/cedr_v1} \\
\midrule
\rowcolor{gray!15} \multicolumn{5}{c}{Intent Classification}\\
Massive Intent~\citep{2023MASSIVE} & 51 & Clustering & 661,923 & \url{huggingface.co/datasets/mteb/amazon_massive_intent} \\
MTOP Intent~\citep{2021MTOP} & 6 & Clustering & 83,922 & \url{huggingface.co/datasets/mteb/mtop_intent} \\
Banking77~\citep{2020Banking77} & en & Clustering & 9,993 & \url{huggingface.co/datasets/mteb/banking77} \\
\midrule
\rowcolor{gray!15} \multicolumn{5}{c}{Domain Classification}\\
Massive Scenario~\citep{2023MASSIVE} & 51 & Clustering & 661,923 & \url{huggingface.co/datasets/mteb/amazon_massive_scenario} \\
MTOP Domain~\citep{2021MTOP} & 6 & Clustering & 83,922 & \url{huggingface.co/datasets/mteb/mtop_domain} \\
\midrule
\rowcolor{gray!15} \multicolumn{5}{c}{Language Classification}\\
BactrianX Language Classification~\citep{2023bactrianx} & 52 & Clustering & 491,405 & \url{huggingface.co/datasets/MBZUAI/Bactrian-X} \\
\midrule
\rowcolor{gray!15} \multicolumn{5}{c}{Citation Prediction}\\
S2ORC-TItle-Citation~\citep{2020S2ORC} & en & Retrieval & 132,879 & \url{huggingface.co/datasets/sentence-transformers/s2orc} \\
S2ORC-Abstract-Citation~\citep{2020S2ORC} & en & Retrieval & 231,587 & \url{huggingface.co/datasets/sentence-transformers/s2orc} \\
SPECTER~\citep{2020SPECTER} & en & Retrieval & 24,717 & \url{huggingface.co/datasets/sentence-transformers/specter} \\
\midrule
\rowcolor{gray!15} \multicolumn{5}{c}{Linguistic Acceptability}\\
MELA~\citep{2024MELA} & 10 & Classification & 40,267 & \url{huggingface.co/datasets/Geralt-Targaryen/MELA} \\
ScaLA~\citep{2023ScaLA} & 9 & Classification & 128,471 & \url{huggingface.co/datasets/alexandrainst/scala} \\
DaLA~\citep{2025DaLA} & da & Classification & 6,508 & \url{huggingface.co/datasets/giannor/dala_large} \\
\midrule
\rowcolor{gray!15} \multicolumn{5}{c}{Claim Verification}\\
FEVER~\citep{2018FEVER} & en & Retrieval & 106,605 & \url{huggingface.co/datasets/mteb/fever} \\
SciFact~\citep{2020SciFact} & en & Retrieval & 859 & \url{huggingface.co/datasets/mteb/scifact} \\
COLIEE~\citep{2022COLIEE} & en & Retrieval & 454 & \url{www.modelscope.cn/datasets/sentence-transformers/coliee} \\
FEVER-NL~\citep{2025beir-nl} & nl & Retrieval & 94,987 & \url{huggingface.co/datasets/clips/beir-nl-fever} \\
\midrule
\rowcolor{gray!15} \multicolumn{5}{c}{STS}\\
STS12~\citep{2012STS12} & en & Retrieval & 1,858 & \url{huggingface.co/datasets/mteb/sts12-sts} \\
STS22~\citep{2022STS22} & en & Retrieval & 389 & \url{huggingface.co/datasets/mteb/sts22-crosslingual-sts} \\
STSBenchmark~\citep{2021STSBenchmark} & en & Retrieval & 3,297 & \url{huggingface.co/datasets/mteb/stsbenchmark-sts} \\
STS22-Crosslingual~\citep{2022STS22} & 7 & Retrieval & 1,469 & \url{huggingface.co/datasets/mteb/sts22-crosslingual-sts} \\
BQ~\citep{2023CMTEB} & zh & Retrieval & 2,436 & \url{huggingface.co/datasets/C-MTEB/BQ} \\
QBQTC~\citep{2021qbqtc} & zh & Retrieval & 37,139 & \url{github.com/CLUEbenchmark/QBQTC} \\
SimCLUE~\citep{2022simclue} & zh & Retrieval & 213,301 & \url{github.com/CLUEbenchmark/SimCLUE} \\
LCSTS~\citep{2015LCSTS} & zh & Retrieval & 278,146 & \url{huggingface.co/datasets/hugcyp/LCSTS} \\
Nordic STS~\citep{2025nordic_sts} & da, sv, no & Retrieval & 70,617 & \url{huggingface.co/datasets/kardosdrur/synthetic-nordic-sts} \\
    \bottomrule
    \end{tabular}
    }
    \caption{Number of samples in our collected training dataset (part 2).}
    \label{tab:data-2}
\end{table}

\clearpage
\section{Details on MTEB Evaluation}\label{appendix:mteb}

The Massive Text Embedding Benchmark (MTEB) is widely recognized as the de facto standard for the comprehensive evaluation of text embedding models. Originally introduced by \citet{2023MTEB}, it was vastly expanded into the Massive Multilingual Text Embedding Benchmark (MMTEB) through a large-scale, open-science collaboration~\citep{2025MMTEB}. This community-driven effort has established a rigorous and diverse evaluation framework, encompassing over 500 quality-controlled tasks that span more than 250 languages and a wide array of domains.

The significance of MTEB lies in its unprecedented scale and diversity, which addresses the critical limitations of previous benchmarks that were often constrained to a few languages (mostly English), specific domains (e.g., news), or a single task type (e.g., retrieval). To provide a holistic assessment of a model's capabilities, MTEB organizes its evaluation tasks into ten distinct categories:
\begin{itemize}
    \item Retrieval: Assesses a model's ability to find relevant documents from a large corpus for a given query.
    \item Reranking: Measures the ability to reorder a given list of candidate documents by their relevance to a query.
    \item Classification: Evaluates performance on standard text classification tasks (e.g., sentiment analysis, topic classification).
    \item Clustering: Tests how well embeddings group semantically similar documents together.
    \item Pair Classification: Involves predicting the relationship between a pair of texts (e.g., paraphrase detection, natural language inference).
    \item Semantic Textual Similarity (STS): Measures the ability to predict the degree of semantic similarity between two sentences on a continuous scale.
    \item Bitext Mining: Assesses the ability to identify translated sentence pairs from a collection of sentences in two languages.
    \item Summarization: Evaluates the semantic similarity between a model-generated summary and a reference summary.
    \item Instruction Reranking: A more challenging reranking variant where the model must follow a detailed natural language instruction to determine relevance.
    \item Multilabel Classification: A classification variant where each document can be assigned multiple labels.
\end{itemize}

The hundreds of tasks are further organized into benchmarks, which are curated subsets of tasks grouped by language, domain, or a combination of both. This includes language-specific benchmarks such as English, Chinese, and Russian; domain-specific benchmarks such as Code and Medical; and aggregated benchmarks like Multilingual, European, and Scandinavian, which test performance across a broad and diverse set of languages. This hierarchical structure allows for both a fine-grained analysis of a model's performance on a specific language or domain and a high-level view of its overall multilingual and multi-domain capabilities.

In this work, we leverage the breadth of MTEB to provide a robust and thorough evaluation of our models. We evaluate on \tb{17 benchmarks}, totaling \tb{430 unique tasks}: Multilingual, Code, Medical, English, Russian, French, German, Polish, Dutch, Indic, Persian, Chinese, Japanese, Korean, Vietnamese, European, and Scandinavian. This extensive evaluation allows for a robust and fine-grained assessment of our models' capabilities, directly supporting our claims of multilingual inclusivity and broad domain competence.

\end{document}